%% file: acl_latex.tex
\pdfoutput=1
\documentclass[11pt]{article}
\usepackage{multido}
\usepackage{amssymb} 
\usepackage{times}
\usepackage{latexsym}
\usepackage{color,soul}
\usepackage[T1]{fontenc}
\usepackage[utf8]{inputenc}
\usepackage[table]{xcolor}
\definecolor{lightblue}{RGB}{212, 244, 252} 
\definecolor{lightyellow}{RGB}{255, 242, 206}
\definecolor{gray}{RGB}{230, 230, 230}
\usepackage{microtype}
\usepackage[final]{acl}
\usepackage{inconsolata}
\usepackage{graphicx}
 \usepackage{booktabs}
\usepackage{graphicx}
\usepackage{xcolor}
\definecolor{mycolor}{HTML}{dbf0ff}
\usepackage{multirow}
\usepackage{booktabs}
\usepackage{algorithm}
\usepackage{algpseudocode}
\usepackage{footnote}
\usepackage{enumitem}
\makesavenoteenv{tabular}
\makesavenoteenv{table}
\usepackage{amsmath}
\usepackage{multirow}
\usepackage{xcolor}
\definecolor{customgreen}{HTML}{48D8B2}
\definecolor{rr}{HTML}{e03c39}
\definecolor{orange}{HTML}{f68d6a}
\definecolor{blueish}{HTML}{557aa0}

\newcommand{\orange}[1]{\textcolor{orange}{#1}}

\title{Train Once for All: A Transitional Approach for \\Efficient Aspect Sentiment Triplet Extraction}
\small{\author{
  Xinmeng Hou$^{1}$, Lingyue Fu$^{2}$, Chenhao Meng$^{2}$, Kounianhua Du$^{2}$, Wuqi Wang$^{3}$, Hai Hu$^{2\ast}$\thanks{$^*$ Corresponding Author} \\
  $^{1}$Columbia University, New York, NY, USA \\
  $^{2}$Shanghai Jiao Tong University, Shanghai, China \\
  $^{3}$Chang’an University, Xi’an 710064, China \\
  fh2450@tc.columbia.edu, \quad fulingyue@sjtu.edu.cn, \quad chenhaomeng@sjtu.edu.cn, \\ kounianhuadu@sjtu.edu.cn, \quad wuqiwang@chd.edu.cn,\quad hu.hai@sjtu.edu.cn}
}

\begin{document}
\maketitle

\begin{abstract}
% Aspect-Opinion Pair Extraction (AOPE) and Aspect Sentiment Triplet Extraction (ASTE) have gained significant attention in the field of natural language processing. However, most existing methods extract aspects/opinions and identify their relations \orange{directly and separately}, leading to error propagation and high time complexity be.
Aspect-Opinion Pair Extraction (AOPE) and Aspect Sentiment Triplet Extraction (ASTE) have drawn growing attention in NLP. However, most existing approaches extract aspects and opinions 
% simultaneously and 
independently, optionally adding pairwise relations, often leading to error propagation and high time complexity.
% HH: 前面这句话的逻辑要理清楚。是因为把ASTE分成两个独立的任务，所以造成了error propagation和high time complexity吗？
% \orange{directly and separately $\rightarrow$ sequentially??}
% \orange{What exactly is the cause for high time complexity? Is high time complexity caused by having two separate subtasks?}
To address these challenges and being inspired by transition-based dependency parsing, we propose the first transition-based model for AOPE and ASTE that performs aspect and opinion extraction jointly, which also better captures position-aware aspect-opinion relations and mitigates entity-level bias.
% HH: 这句话和前面的接不上。前面提出现有研究有问题A，那我们提出的方案就应该解决问题A，而不是我们的方案解决了问题B。
% 可以改成：To address these challenges, we propose a transition-based model that performs Aspect-Extraction and Opinion-Extraction jointly, which better captures position-aware aspect-opinion relations and mitigates token-level bias.
% 我没懂mitigate token-level bias是什么意思
By integrating contrastive-augmented optimization, our model delivers more accurate action predictions and jointly optimizes separate subtasks in linear time. 
Extensive experiments on 4 commonly used ASTE/AOPE datasets show that, while performing worse when trained on a single dataset than some previous models, our model achieves the best performance on both ASTE and AOPE if trained on combined datasets, outperforming the strongest previous models in F1-measures (often by a large margin). 
% at least 6.98\% in F1. 
We hypothesize that this is due to our model's ability to learn transition actions from multiple datasets and domains.
Our code is available at \url{https://anonymous.4open.science/r/trans_aste-8FCF}.
\end{abstract}

\input{new-intro}

\begin{table*}[t]
\centering
\resizebox{\linewidth}{!}{
\begin{tabular}{l l l}
\toprule
\textbf{Method} & \textbf{Approach} & \textbf{Time Complexity} \\
\midrule
Peng-Two-stage~\citep{Peng_Xu_Bing_Huang_Lu_Si_2020} & Two-Stage Pipeline: entity identification and relation formation & $O(n + k^2)$ \\
BARTABSA~\citep{yan-etal-2021-unified} & Generative-based Aspect-based Sentiment Analysis & $O(m \cdot v)$ \\
GTS~\citep{wu2020grid} & Grid Matrix-based Tagging & $O(n^2)$ \\
JET-BERT~\cite{xu2020position} & Position-Aware Sequence Tagging & $O(n)$ \\
COM-MRC~\cite{zhai2022mrc} & Compositional Machine Reading Comprehension & $O(r \cdot n^2 \cdot h)$ \\
Triple-MRC~\cite{zou2024multi} & Multi-turn Machine Reading Comprehension & $O(r \cdot n^2 \cdot h)$ \\
EMC-GCN~\cite{chen2022enhanced} & Multi-channel Graph Convolutional Network & $O(m \cdot n^2 \cdot h)$ \\
DGCNAP~\cite{li2023dual} & Graph Convolutional Network w/ Affective Knowledge & $O(m \cdot n^2 \cdot h)$ \\
MiniConGTS~\cite{sun-etal-2024-minicongts} & Lightweight Grid Matrix-based Tagging System & $O(n^2)$ \\\midrule
\textbf{Trans-model} (Ours) & Transition-based Action Prediction for Simulating Relation Formation and Pair Extraction & $O(n)$ \\
\bottomrule
\end{tabular}}
\caption{An overview of previous methods and models (which will serve as  baselines in this study), their approaches, and corresponding time complexities. Here, the hidden size for LSTM $d$ is simplified; $n$ is the sequence length; $m$ is the number of graph channels; $v$ is the vocabulary size; $k$ is the number of extracted terms; $r$ is the number of query rounds, and $h$ is the hidden size of the encoder.}
\label{tab:baselines}
\end{table*}

\section{Related Work}

\paragraph{Previous methods on ASTE}
Pipeline-based approaches, such as Peng-Two-stage \cite{Peng_Xu_Bing_Huang_Lu_Si_2020}, decompose the task into multiple stages for modular refinement. Sequence-to-sequence frameworks like BARTABSA \cite{yan-etal-2021-unified} employ pretrained transformers to generate triplets flexibly. Sequence-tagging methods, including GTS \cite{wu2020grid} and JET-BERT \cite{xu2020position}, annotate tokens for precise identification of relationships. Machine Reading Comprehension (MRC)-based models, such as COM-MRC \cite{zhai2022mrc} and Triple-MRC \cite{zou2024multi}, reframe the task as query answering for efficient extraction. Graph-based approaches such as EMC-GCN \cite{chen2022enhanced}, BDTF \cite{chen2022enhanced}, and DGCNAP \cite{li2023dual} use graph structures to capture semantic and syntactic interactions. Tagging schema-based models, exemplified by STAGE-3D \cite{liang2023stage}, use hierarchical schemas for multi-level extraction, while lightweight models like MiniConGTS \cite{sun-etal-2024-minicongts} focus on efficiency with reduced computational costs.

Table~\ref{tab:baselines} summarizes these baseline methods, along with our proposed model, in terms of their core approaches and time complexities.

\paragraph{Transition-based Methods in NLP}
Transition-based approaches are widely used in dependency parsing, leveraging shift-reduce and bidirectional arc actions (left-arc, right-arc) for efficient \(O(n)\)  parsing \cite{aho1973theory, nivre-2003-efficient, cer2010parsing}. These parsers maintain stack, buffer, and arc relations to track transitions and then build up dependency relations between tokens.
% \orange{and then build up dependency relations between tokens.}

Transition-based methods have also been applied to various NLP tasks, including token segmentation \cite{zhang2016transition}, argument mining \cite{bao2021neural}, constituency parsing \cite{yang2020stronglyincrementalconstituencyparsing}, AMR parsing \cite{zhou2021structureawarefinetuningsequencetosequencetransformers}, and sequence labeling \cite{gómezrodríguez2020unifyingtheorytransitionbasedsequence}, among others. Transition-based methods have been explored in emotion analysis \cite{fan2020transition, jian2024emotrans}. In sentiment analysis, however, transition-based models have not been widely adopted. One exception is their use in generating graph structures for opinion extraction \cite{fernándezgonzález2023structuredsentimentanalysistransitionbased}, although this design relies on graph embeddings and thus results in a time complexity of $O(N^2)$, with performance that lags behind more recent AOPE and ASTE approaches.

% and have been extended to tasks like token segmentation \cite{zhang2016transition} and argument mining \cite{bao2021neural}. Beyond parsing, transition-based methods have been applied to structured sentiment analysis, generating graph structures for opinion extraction \cite{fernándezgonzález2023structuredsentimentanalysistransitionbased} and incremental constituency parsing using graph neural networks \cite{yang2020stronglyincrementalconstituencyparsing}. A unifying theory bridges transition-based parsing and sequence labeling for simplified learning \cite{gómezrodríguez2020unifyingtheorytransitionbasedsequence}, while in AMR parsing, structure-aware fine-tuning achieves state-of-the-art results \cite{zhou2021structureawarefinetuningsequencetosequencetransformers}. Their flexibility and efficiency make them effective across diverse NLP tasks.

\paragraph{Contrastive-based Optimization}
Contrastive learning has demonstrated its effectiveness in various domains, achieving state-of-the-art results in token-independent extraction tasks with models such as MiniconGTS \cite{sun-etal-2024-minicongts}. Recent works have further explored its potential: contrastive learning for prompt-based few-shot language learners enhances generalization through augmented ``view''~\cite{jian2022contrastivelearningpromptbasedfewshot}; contrastive learning as goal-conditioned reinforcement learning provides a novel loss function with theoretical guarantees for improved success in goal-directed tasks~\cite{eysenbach2023contrastivelearninggoalconditionedreinforcement}, and studies analyzing the role of margins reveal critical factors like positive sample emphasis for better generalization \cite{rho2023understandingcontrastivelearninglens}. These advances showcase contrastive learning’s flexibility and robustness, which inspired us to integrate it into our transition-based AOPE and ASTE to refine representation learning and improve model performance.

\begin{table*}[h!]
\centering
\small{
\begin{tabular}{ll}
\hline
\textbf{Action} & \textbf{Symbolic Expression} \\ \hline

\textbf{Shift} ($SF$) & 
$(\sigma_0,\, \beta_0 \mid \beta_1, A, O, R) \xrightarrow{SH} (\sigma_0 \mid \sigma_1, \beta_1, A, O, R)$ \\ \hline

\textbf{Stop} ($ST$) & 
$(\sigma_0, , A, O, R) \xrightarrow{ST} (, , A, O, R)$ \\ \hline

\textbf{Merge} ($M$) & 
$(\sigma_0 \mid \sigma_1,\, \beta_1 \mid \beta_2, A, O, R) \xrightarrow{M} (\sigma_{0 \& 1},\, \beta_1 \mid \beta_2, A, O, R)$ \\ \hline

\textbf{Left Constituent Removal} ($L_n$) & 
$(\sigma_0 \mid \sigma_1,\, \beta_0, A, O, R) \xrightarrow{L_n} (\sigma_1,\, \beta_0, A, O, R)$ \\ \hline

\textbf{Right Constituent Removal} ($R_n$) & 
$(\sigma_0 \mid \sigma_1,\, \beta_0, A, O, R) \xrightarrow{R_n} (\sigma_0,\, \beta_0, A, O, R)$ \\ \hline

\textbf{Left-Relation Formation} ($LR$) & 
$(\sigma_0 \mid \sigma_1, \beta_0, A, O, R) \xrightarrow{LR} (\sigma_0 \mid \sigma_1, \beta_0, A \cup \sigma_1, O \cup \sigma_0, R \cup \sigma_0 \xleftarrow{} \sigma_1)$ \\ \hline

\textbf{Right-Relation Formation} ($RR$) & 
$(\sigma_0 \mid \sigma_1, \beta_0, A, O, R) \xrightarrow{RR} (\sigma_0 \mid \sigma_1, \beta_0, A \cup \sigma_0, O \cup \sigma_1, R \cup \sigma_0 \xrightarrow{} \sigma_1)$ \\ \hline

\end{tabular}}
\caption{Symbolic Expressions for the Proposed Actions. Here, $\sigma$ represents the stack, $\beta$ represents the buffer, $A$ denotes the aspect, $O$ denotes the opinion, and $R$ consist of an aspect and an opinion.}
% \orange{HH: what are $\sigma, \beta$...?}
\label{tab:symbolic_actions}
\end{table*}

\section{The Trans-AOPE/ASTE Model}

% In this section, we introduce an innovative paradigm dedicated to incorporating the extraction of aspect-opinion pairs into a process akin to the construction of a parsing-directed graph. The overall framework is presented in Figure \ref{f}. This enhanced model progressively builds and annotates aspect-oriented relations using input sequences embedded with contextual features. As a departure from the traditional approach of employing three sets to document action records, our methodology utilizes five distinct sets. These sets are specifically assigned to record the state pertaining to the stack, buffer, aspect, opinion, and edge representing the corresponding relation. The subsequent section will delve into the system's computational algorithms designed to optimize the performance of triplet extraction and provide a comprehensive exploration of the underlying methodologies and techniques used to efficiently and accurately identify and extract aspect-sentiment triplets.

% \orange{This paragraph should give a very high-level description of the algorithm.}
In this section, we introduce a novel paradigm that treats the extraction of aspect-opinion pairs as a parsing-directed graph construction process. This paradigm comprises two major procedures: aspect-opinion pair extraction and tagging. We refer to the pair-extraction model as \emph{Trans-AOPE} and the model that incorporates tagging as \emph{Trans-ASTE}. Trans-AOPE progressively constructs and annotates aspect-oriented relations using input sequences enriched with contextual features. Unlike classic approaches, which employ three sets to record action histories, our paradigm utilizes five sets (including a stack, buffer, aspect set, and pair set) to demonstrate how aspects and opinions are extracted together.

% \orange{We name our method Trans-AOPE or Trans-ASTE depending on the specific task.}
% \fly{What five sets are?}

\subsection{Transitional Operations and State Change}
\begin{table*}[h!]
\centering
\small{
\begin{tabular}{ccllccc}
\hline
\textbf{Phrase} & \textbf{Action} & \textbf{Stack ($\sigma$)} & \textbf{Buffer ($\beta$)} & \textbf{Aspect} & \textbf{Opinion} & \textbf{Pair} \\ \hline

-- & -- & [] & [$\beta_1$, $\beta_2$, $\beta_3$, $\beta_4$] & -- & -- & -- \\ \hline

1 & $SF$ & [$\sigma_1$] & [$\beta_2$, $\beta_3$, $\beta_4$] & -- & -- & -- \\ \hline

2 & $SF$ & [$\sigma_1$, $\sigma_2$] & [$\beta_3$, $\beta_4$] & -- & -- & -- \\ \hline

3 & $M$ & [$\sigma_{1\&2}$] & [$\beta_3$, $\beta_4$] & -- & -- & -- \\ \hline

4 & $SF$ & [$\sigma_{1\&2}$, $\sigma_3$] & [$\beta_4$] & -- & -- & -- \\ \hline

5 & $R_n$ & [$\sigma_{1\&2}$] & [$\beta_4$] & -- & -- & -- \\ \hline

6 & $SF$ & [$\sigma_{1\&2}$, $\sigma_4$] & [] & -- & -- & -- \\ \hline

7 & $RR$ & [$\sigma_{1\&2}$, $\sigma_4$] & [] & [$\sigma_{1\&2}$] & [$\sigma_4$] & ($\sigma_{1\&2} \rightarrow \sigma_4$) \\ \hline

9 & $ST$ & [] & [] & [$\sigma_{1\&2}$] & [$\sigma_4$] & ($\sigma_{1\&2} \rightarrow \sigma_4$) \\ \hline

\end{tabular}}
\caption{State changes for "Gourmet food is delicious" using symbolic representation. Here, $\sigma_1$ corresponds to "Gourmet", $\sigma_2$ to "food", $\sigma_3$ to "is", and $\sigma_4$ to "delicious". Similarly, $\beta_1$, $\beta_2$, $\beta_3$, and $\beta_4$ correspond to tokens in the buffer in sequence.}

\label{tab:state_changes1}
\end{table*}

The construction of relationships between phrases is achieved through systematic transition actions, conceptualized as directed edges linking two nodes (tokens), \(N_1\) and \(N_2\). A relationship transitioning \(N_1\) to \(N_2\) under relation \(l_1\) is denoted as \(RR = N_1 \xrightarrow{l_1} N_2\) or \(LR = N_1 \xleftarrow{l_1} N_2\), representing aspect-oriented direction. To interpret causal links, \(l_1\) is defined as \(l_L\) or \(l_R\) for bidirectional relations. Opinion constituents (\(E_1\)) may precede or follow aspect constituents, which can span multiple tokens (e.g., ``gourmet food,'' ``not bad''). This requires additional steps to consolidate multi-token constituents. For the ASTE
% \fly{use ASTE is ok, you have defined before.
task, we define seven distinct transition actions that combine token retrieval, termination, and token merging.

The action framework aims to prevent information leakage and ensure flexibility in relation arching. Actions are recorded as tuples $T = (\sigma, \beta, A, O, R)$, representing the stack, buffer, aspect, opinion, and relation. The ASTE task includes default actions, which are universal and required for all data, primary actions that handle merging or removal, and secondary actions for relation arching, which depend on the outcomes of the primary actions. Verbal and symbolic representations of state transitions for each action are provided below.

\paragraph{Default Actions:}

\begin{enumerate}[topsep=2pt, itemsep=0pt, parsep=2pt]
    \item \textit{\textbf{Shift}} ($SF$) moves a token from the tokenized stack into the buffer for further processing.
    \item \textit{\textbf{Stop}} ($ST$) halts the process when only one token remains in the buffer, and the stack is empty.
\end{enumerate}

\paragraph{Primary Actions:}

\begin{enumerate}[topsep=2pt, itemsep=0pt, parsep=2pt]
    \item \textit{\textbf{Merge}} ($M$) combines multiple tokens in the buffer into a single compound target.
    \item \textit{\textbf{Left Constituent Removal}} ($L_n$) removes the left constituent from the buffer.
    \item \textit{\textbf{Right Constituent Removal}} ($R_n$) removes the right constituent from the buffer.
\end{enumerate}

\paragraph{Secondary Actions:}

\begin{enumerate}[topsep=2pt, itemsep=0pt, parsep=2pt]
    \item \textit{\textbf{Left-Relation Formation}} ($LR$) creates a relation from the right aspect constituent to the left opinion constituent.
    \item \textit{\textbf{Right-Relation Formation}} ($RR$) creates a relation from the left aspect constituent to the right opinion constituent.
\end{enumerate}

Table \ref{tab:symbolic_actions} provides a symbolic illustration of how the symbolic state is constructed and utilized. Take the sentence "Gourmet food is delicious" as an example. Table \ref{tab:state_changes1} demonstrates the process of moving tokens from the buffer to the stack, deciding whether they should be merged into a single entity or removed, and finally evaluating them for relation formation. It is important to note that the set of actions shown in the figure is not the only way to extract the "Gourmet food" and "delicious" aspect-opinion pair. An alternative approach use the stack's capacity of  holding multiple tokens, moving "is" to the stack ($\beta_3 \rightarrow \sigma_3$) before merging "Gourmet" and "food" ({$[\sigma_1, \sigma_2, \sigma_3] \rightarrow [\sigma_{1\&2}, \sigma_3]$}).

\subsection{Trans-AOPE State Representation}
\begin{figure}[tb!]
\begin{center}
\includegraphics[width=0.95\linewidth]{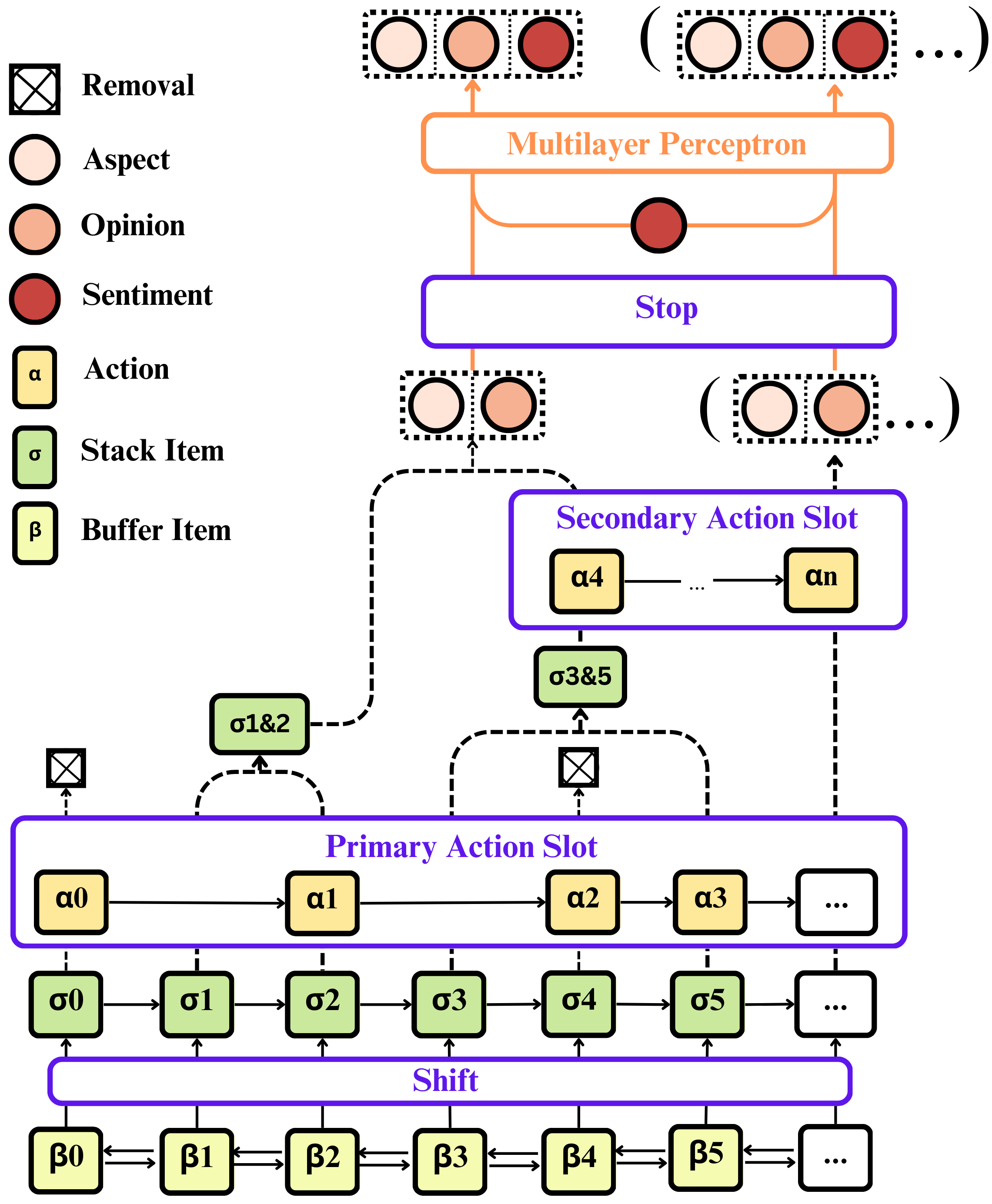}
\caption{The complete process of the transition-based model is illustrated. Purple highlights represent the transition-based pair extraction, while orange indicates the final step of sentiment tagging.
}
\label{framework}
\end{center}
\end{figure}
% In this model, the input, which is indicated as $I^n_1=(t_1, t_2, \ldots, t_n)$, consists of a series of tokens. Conversely, the output is a sequence of actions denoted as $A^m_1=(a_1, a_2, \ldots, a_m)$. This scenario can essentially be viewed as a search operation for an optimal action sequence, $A^*$, given a sequence of clauses, $I^n_1$. At any given step, specifically step $n$, our model predicts the next action by relying on the current system state, $S$, and a series of prior actions, $A^{n-1}_1$. Concurrently, $S_{n+1}$ is the system state, updated in accordance with the specific action, $a_t$. We introduce $r_n$ as a symbolic representation utilized for calculating the probability of the action $a_n$ at step $n$. This concept can be mathematically expressed as:
% \fly{Give the model/pipeline a name for clarity. Also, each stage needs a name to substitute first stage and second stage.}
The model we propose consists of two core stages: pair extraction with a designed transitional action slot (in purple) and pair-based sentiment tagging (in orange), as illustrated in Figure \ref{framework}.

In the first stage, the input, denoted as $I^n_1 = (t_1, t_2, \ldots, t_n)$, is a sequence of tokens. The output is a sequence of actions, represented as $A^m_1 = (a_1, a_2, \ldots, a_m)$. This process can be conceptualized as a search for the optimal action sequence, $A^*$, given the input sequence $I^n_1$. At each step $n$, the model predicts the next action based on the current system state, $S$, and the sequence of prior actions, $A^{n-1}_1$
% \fly{$A^{n-1}_1$?}. 
The updated system state, $S_{n+1}$
% \fly{$S_{n+1}$?}
, is determined by the specific action $a_t$. We define $r_n$ as a symbolic representation for calculating the probability of the action $a_n$ at step $n$. This probability is computed as follows:

\begin{equation} p(a_n|r_n) = \frac{\exp(w_{a_n}^{\top}r_n + b_{a_n})}{\sum\nolimits_{a' \in \mathcal{A}(S)}{\exp(w_{a'}^{\top}r_n + b_{a'})}} \end{equation}
% \fly{No need for a new paragraph}
Here, $w_a$ is a learnable parameter vector, and $b_a$ is a bias term. The set $\mathcal{A}(S)$ represents the legal actions available given the current parser state. The overall optimization objective for the model is defined as:

\begin{equation}
\label{likelihood}
\begin{aligned}
    (A^*, S^*) &= \underset{_{A,S}}{\mathrm{argmax}}\prod_{n} p(a_n, S_{n+1}|A^{n-1}_1, S_{n}) \\
               &= \underset{_{A,S}}{\mathrm{argmax}}\prod_{n} p(a_n|r_n)
\end{aligned}
\end{equation}
In this stage, the ASTE task is incorporated into a transition-based action prediction task. In order to decode efficiently, the action boasting the maximum probability is chosen through a greedy approach until the parsing procedure reaches a point of termination. This model thus forms the basis of an efficient transition-based prediction system that utilizes a representation of the current system state and action history to generate the most likely subsequent action at each step in a sequence. By integrating these concepts, the model successfully carries out parsing tasks while avoiding information leakage and ensuring flexibility in the relation arching process.

\subsection{Transition Implementation with Neural Model}
This section introduces a transition-based parsing process. RoBERTa \cite{liu2019robertarobustlyoptimizedbert} encodes the text, while UniLSTM \cite{10.1162/neco.1997.9.8.1735} and BiLSTM \cite{graves2005framewise} capture transitions. The parser state evolves through a sequence of actions, with LSTMs processing each token once. This yields a time complexity of \(O(n \cdot d^2)\), typically simplified to \(O(n)\) under fixed \(d\). Finally, an MLP classifies the sentiment for each pair or triplet based on the final parser state.

% \fly{I think $\setminus$paragraph is better than $\setminus$subsubsection.}
\paragraph{Token representations}  
Consider the process of parsing a text $d^n_1$ = ($p_1$, $p_2$, $\ldots$, $p_n$) 
% \fly{text or sentence? proper nouns should be unified in the full paper.}
, consisting of $n$ phrases. Each phrase $p_i$ = ($w_{i1}$, $w_{i2}$, $\ldots$, $w_{il}$) contains $l$ tokens. A phrase can be represented as a sequence $x_i$ = ($\mathrm{[CLS]}$, $t_{i1}$, $\ldots$, $t_{il}$, $\mathrm{[SEP]}$), where $\mathrm{[CLS]}$ is a special classification token whose final hidden state serves as the aggregate sequence feature, and $\mathrm{[SEP]}$ is a separator token. The hidden representation of each phrase is computed as $h_{p_i}$ = $\mathrm{RoBERTa}$($x_i$) $\in \mathbb{R}^{d_b \times |l_i|}$, where $d_b$ is the hidden dimension size, and $|l_i|$ is the length of the sequence $x_i$. Finally, the entire text $d^n_1$ is represented as a list of tokens: $h_d$ = [$h_{p_1}$, $h_{p_2}$, $\ldots$, $h_{p_n}$].
% \fly{what is [] mean? concatentation or list?}.

\paragraph{State Initialization} 
 At the start of the parsing process, the parser's state is initialized as $(\beta = \varnothing, \sigma = [1, 2, \ldots, n], E = \varnothing, C = \varnothing, R = \varnothing)$, where $\sigma$ is the stack, $\beta$ is the buffer, and $E$, $C$, and $R$ are empty sets representing different outputs. The state evolves through a sequence of actions, progressively consuming elements from the buffer $\beta$ and constructing the output. This process continues until the parser reaches its terminal state when there is only one token left in buffer, represented as $(\beta = [SEP], \sigma = \varnothing, E, C, R)$.
 % \fly{What is \$ mean?}

\paragraph{Step-by-Step Parser State Representation.} 
For the action sequence, each action $a$ is mapped to a distributed representation $e_a$ through a lookup table $E_a$. An unidirectional LSTM is then utilized to capture the complete history of actions in a left-to-right manner at each step $t$:
\begin{equation}
\begin{aligned}
    \alpha_t = \text{LSTM}_a(a_{0}, a_{1}, \ldots, a_{t-1}, a_{t})
\end{aligned}
\end{equation}
Upon generation of a new action $a_t$, its corresponding embedding $e_{a_t}$ is integrated into the rightmost position of LSTM$_a$. To further refine the representation of the pair $(\sigma_1, \sigma_0)$, their relative positional distance $d$ is also encoded as an embedding $e_d$ from a lookup table $E_d$. The composite representation of the parser state at step $t$ encompasses these varied features.

The parser state is represented as a triple $(\beta_s, \sigma_s, A_t)$, where $\sigma_s$ denotes the stack sequence $(\sigma_0, \sigma_1, \ldots, \sigma_n)$, $\beta_s$ represents the buffer sequence $(\beta_0, \beta_1, \ldots, \beta_n)$, and $A_t$ encapsulates the action history $(a_0, a_1, \ldots, a_{t-1}, a_t)$. The stack ($\sigma_n$) and buffer ($\beta_n$) are encoded using bidirectional LSTMs as follows:
\begin{equation}
\begin{aligned}
[s_t, b_t] = \text{BiLSTM}((\sigma_n, \beta_0), (\sigma_{n-1}, \beta_1), \\ \ldots, (\sigma_0, \beta_n))
\end{aligned}
\end{equation}
% \fly{No need new paragraph} 
Here, $s_t$ and $b_t$ are the output feature representations of the stack and buffer, respectively. Each of these representations consists of forward and backward components: $\sigma_t = (\overrightarrow{\sigma_t}, \overleftarrow{\sigma_t})$ and $\beta_t = (\overrightarrow{\beta_t}, \overleftarrow{\beta_t})$. The forward and backward components are matrices in $\mathbb{R}^{d_l \times |\sigma_t|}$ and $\mathbb{R}^{d_l \times |\beta_t|}$, respectively, where $d_l$ is the hidden dimension size of the LSTM, and $|\sigma_t|$, $|\beta_t|$ are the lengths of the sequences $\sigma_t$ and $\beta_t$. 
% \fly{What is the use of $s_t$ and $b_t$?}

% \begin{figure*}[h!]
%   \centering
%   \begin{minipage}[b]{0.45\linewidth}
%     \centering
%     \includegraphics[width=\linewidth]{15res.png}
%     \caption{F1 scores on ASTE task over the training epochs for the model trained on 15res and evaluated on 14res, 14lap, 15res, and 16res datasets. The highest F1 for 15res is 64.04\%, while for 16res it is 74.08\%.}
%     \label{fig:15res}
%   \end{minipage}
%   \hfill
%   \begin{minipage}[b]{0.45\linewidth}
%     \centering
%     \includegraphics[width=\linewidth]{16res.png}
%     \caption{F1 scores on ASTE task over the training epochs for the model trained on 16res and evaluated on 14res, 14lap, 15res, and 16res datasets. The highest F1 for 15res is 82.07\%, while for 16res it is 61.89\%.}
%     \label{fig:16res}
%   \end{minipage}
% \end{figure*}

\subsection{Optimization Implementation}
We compare two optimization strategies: regular optimization using Cross-Entropy Loss and contrastive-based optimization, which aligns predicted and true action embeddings. A weight study will evaluate the impact of the positioning of two components in augmented optimization on model performance.
% \fly{Say two variants, do not call it ablation study. You can write these two variants focus on different aspects for different scenarios.}
\subsubsection{Base Optimization} 
Both action and sentiment classification tasks are optimized using the Cross-Entropy Loss as base optimization, defined as follows:
\begin{equation}
\mathcal{L} = - \frac{1}{N} \sum_{i=1}^{N} \sum_{j=1}^{M} y_i^j \log(p_i^j),
\end{equation}
where \(N\) is the number of samples, \(M\) is the number of classes (either action or sentiment classes), \(y_i^j\) is a binary indicator (0 or 1) indicating whether class \(j\) is the correct class for sample \(i\), and \(p_i^j\) is the predicted probability for class \(j\) for sample \(i\). For AOPE task, total loss is action loss $\mathcal{L}_{\text{action}}$, and for ASTE task, is the sum of the losses for both tasks: $\mathcal{L}_{\text{base}} = \mathcal{L}_{\text{action}} + \mathcal{L}_{\text{sentiment}}.$
% \fly{sum for real number is + rather than $\oplus$}

\subsubsection{Contrastive-based Optimization} 
Given the predicted action logits \(\mathbf{A}_{\text{logits}}\) and the true action labels \(\mathbf{A}_{\text{true}}\), the predicted actions are obtained by applying a softmax function followed by an \(\text{argmax}\): 
\[
\mathbf{A}_{\text{pred}} = \text{argmax}(\text{softmax}(\mathbf{A}_{\text{logits}}))
\]

The predicted and true actions are then embedded as follows:  
\[
\mathbf{E}_{\text{pred}} = \text{Embed}(\mathbf{A}_{\text{pred}}), \quad \mathbf{E}_{\text{true}} = \text{Embed}(\mathbf{A}_{\text{true}})
\]

For each sample \(i\), compute the cosine similarity between \(\mathbf{E}_{\text{pred}}^{(i)}\) and all \(\mathbf{E}_{\text{true}}^{(j)}\) to form a similarity matrix. Using this matrix, the diagonal elements represent positive pairs, while the non-diagonal elements represent negative pairs.

Next, calculate the positive and negative cosine similarities, and apply the exponential function:
\begin{equation}
    e_{\text{pos}} = \exp(\text{cos\_sim}(\mathbf{E}_{\text{pred}}, \mathbf{E}_{\text{true}}) \cdot \mathbf{M}_{\text{pos}})
\end{equation}
\begin{equation}
    e_{\text{all}} = \exp(\text{cos\_sim}(\mathbf{E}_{\text{pred}}, \mathbf{E}_{\text{true}}))
\end{equation}
% \fly{no need new paragraph. Explain cos\_sim means cos similarity function. $Exp_pos$ is not a good name, maybe $e_{pos}$ is better.}
where, cos\_sim denote cosine similarity function \(\mathbf{M}_{\text{pos}}\) is a mask that selects only positive pairs (diagonal entries).

Now compute the contrastive loss as:
\begin{equation}
\mathcal{L}_{\text{con}} = -\frac{1}{N} \sum_{j=1}^{N} \log \left( \frac{\exp_{\text{pos}}^{(j)}}{\exp_{\text{all}}^{(j)}} \right)
\end{equation}
% \fly{$(i)->i$.}
% \fly{similarly, $\oplus \rightarrow +$.}
Finally, the total loss is computed as:
\begin{equation}
\mathcal{L}_{\text{total}} =  \omega_1 \cdot \mathcal{L}_{\text{base}} + \omega_2 \cdot \mathcal{L}_{\text{con}} 
\label{10}
\end{equation}
where \(\omega\) are weighting factors, and \(\mathcal{L}_{\text{base}}\) represents any regularization loss added to improve model generalization.

\section{Experimental Setups}
\subsection{Datasets and preprocessing}
We use well-established benchmarks in Sentiment Analysis: \textbf{14lap}, \textbf{14res} from SemEval-2014~\cite{pontiki-etal-2014-semeval}, \textbf{15res} from SemEval-2015~\cite{pontiki-etal-2015-semeval}, and \textbf{16res} from SemEval-2016~\cite{pontiki-etal-2016-semeval} (see Table~\ref{tab:dataset:stats}).
The first is a collection of \textbf{lap}top reviews while the other three are curated from \textbf{res}taurant reviews. 
All are widely used for aspect-based sentiment analysis and extraction tasks \cite{xu2021positionaware}. Following previous literature, the four datasets are split in a 7:1.5:1.5 proportion for training, development, and test.

\begin{table}[h!]
\centering
\resizebox{\linewidth}{!}{
\begin{tabular}{cccccc}
\toprule
\textbf{Datasets} & \textbf{\#S} & \textbf{\#POS} & \textbf{\#NEU} & \textbf{\#NEG} & \textbf{\#T} \\
\midrule
\textbf{14res} & 2068 & 2869 & 286 & 754 & 3909 \\
\midrule
\textbf{14lap} & 1453 & 1350 & 225 & 774 & 2349 \\
\midrule
\textbf{15res} & 1075 & 1285 & 61 & 401 & 1747 \\
\midrule
\textbf{16res} & 1393 & 1674 & 90 & 483 & 2247 \\
\bottomrule
\end{tabular}}
\caption{Statistics of four datasets. \#S denotes the number data pieces, \#POS, \#NEU, \#NEG the number of positive, neutral and negative sentiment labels, and \#T the total number of triplets.}
\label{tab:dataset:stats}
\end{table}

% SemEval 2014 (\textbf{14lap}, \textbf{14res}) \cite{pontiki-etal-2014-semeval}, SemEval 2015 (\textbf{15res}) \cite{pontiki-etal-2015-semeval}, and SemEval 2016 (\textbf{16res}) \cite{pontiki-etal-2016-semeval}. These datasets, originally curated for the Semantic Evaluation (SemEval) competitions, are widely used for aspect-based sentiment analysis and extraction tasks \cite{xu2021positionaware}. Datasets are split in a 7:1.5:1.5 proportion for training, development, and test sets, and the general statistics are listed in Appendix \ref{data_stats}.

A data processing framework is used to construct sentiment-enhanced dependency graphs for ASTE datasets. This framework utilizes SpaCy \cite{spacy2020} to extract syntactic dependencies and integrates sentiment scores from SenticNet \cite{cambria2017senticnet} to generate weighted adjacency matrices. These enriched graph representations aim to expand the feature space and reduce biased learning. Along with tokenized sentences and their corresponding aspect-opinion-sentiment triplets, the processed data supports downstream tasks, including model training and evaluation.

\subsection{Training settings}
We experiment with two settings for  training data.

\paragraph{In-domain.} In this setting, we train with one of the four datasets and test on the test set of the \textit{same} dataset, e.g., train on 14lap and test on 14lap.

\paragraph{Combined-train.}
Since our method depends on the model learning the correct action to pair an aspect with an opinion from the training data, we hypothesize that it will have the advantage of being able to make use of training data from diverse domains to learn various actions. 
Thus we experiment with training on \textit{two or more training sets combined}, to observe whether there is performance gain when more actions are learned.
Specifically, we train on several training sets together, and evaluate on a single test set.

% We evaluated the action coverage of our transition operations in extracting aspects and opinions, with action evaluation results listed in Appendix \ref{data_conformity}. Since our approach predicts actions rather than tokens, we assessed how well the model learns action patterns across individual datasets. For instance, as shown in Figures \ref{fig:15res} and \ref{fig:16res}, the model did not achieve its highest performance on the corresponding test set. This suggests that the 16res dataset may provide richer action patterns that help the model perform better on 15res, while the 15res dataset may contain patterns that improve performance on 16res. Consequently, we incorporate \textit{OmniTrain}, our multi-dataset training strategy
% \orange{Consequently, we incorporate \textit{OmniTrain}, our multi-dataset training strategy}
% , to provide a more diverse context. This helps the model learn these actions more effectively, which we then evaluate on separate test sets.

\subsection{Baselines}
We compare our models with two groups of baselines: (1) published results from prior studies that do not have runnable code available and (2) previous 
% \orange{SOTA models re-trained by us using OmniTrain.} 
State-of-the-Art (SOTA) models re-trained by us using publicly available code base.

Group (1), as summarized in Table~\ref{tab:baselines}, include CMLA~\cite{wang2017coupled}, Peng-Two-stage~\cite{Peng_Xu_Bing_Huang_Lu_Si_2020}, BARTABSA~\cite{yan-etal-2021-unified}, GTS~\cite{wu2020grid}, JET-BERT~\cite{xu2020position}, COM-MRC~\cite{zhai2022mrc}, Triple-MRC~\cite{zou2024multi}, EMC-GCN~\cite{chen2022enhanced}, DGCNAP~\cite{li2023dual}, and the latest MiniConGTS~\cite{sun-etal-2024-minicongts}. 
% These models are summarized in Table~\ref{tab:baselines}.

For re-trained models in group (2), we use two state-of-the-art models with publicly available and runnable code: the most recent model, \textit{MiniConGTS}~\cite{sun-etal-2024-minicongts}\footnote{\orange{\url{https://github.com/qiaosun22/MiniConGTS}}}, which achieves SOTA performance on both ASTE and AOPE tasks with a complexity of $O(n^2)$ (fixed constants are omitted for simplicity), and an earlier study, \textit{BARTABSA}~\cite{yan-etal-2021-unified}\footnote{\orange{\url{https://github.com/yhcc/BARTABSA}}}, designed for the AOPE task with a complexity of $O(n)$. We use their original configurations and data processing pipelines, modifying only the training data to include all four training sets from 14lap, 14res, 15res, and 16res, when necessary. 
The models are then separately evaluated on the four test sets.

\begin{table*}[h!]
\centering
\resizebox{\textwidth}{!}{
\begin{tabular}{l|ccc|ccc|ccc|ccc}
\toprule
 & \multicolumn{3}{c}{\textbf{14res}} & \multicolumn{3}{c}{\textbf{14lap}} & \multicolumn{3}{c}{\textbf{15res}} & \multicolumn{3}{c}{\textbf{16res}} \\
\cmidrule(lr){2-4} \cmidrule(lr){5-7} \cmidrule(lr){8-10} \cmidrule(lr){11-13}
 \textbf{In-domain setting} & P & R & F1 & P & R & F1 & P & R & F1 & P & R & F1 \\
\midrule
\midrule
CMLA+\cite{wang2017coupled} $\diamond$ & - & - & 48.95 & - & - & 44.10 & - & - & 44.60 & - & - & 50.00 \\
Peng-two-stage \cite{Peng_Xu_Bing_Huang_Lu_Si_2020} $\diamond$ & - & - & 56.10 & - & - & 53.85 & - & - & 56.23 & - & - & 60.04 \\
Dual-MRC \cite{mao2021joint} $\diamond$ & - & - & 74.93 & - & - & 63.37 & - & - & 64.97 & - & - & 75.71 \\
SpanMlt \cite{zhao-etal-2020-spanmlt} $\bullet$ & - & - & 75.60 & - & - & 68.66 & - & - & 64.68 & - & - & 71.78 \\
BARTABSA \cite{yan-etal-2021-unified} $\diamond$ & - & - & 77.68 & - & - & 66.11 & - & - & 67.98 & - & - & \underline{77.38} \\
MiniConGTS \cite{sun-etal-2024-minicongts} $\star$ & - & - & \underline{79.60} & - & - & \underline{73.23} & - & - & \underline{73.87} & - & - & 76.29 \\
\midrule
% \textbf{Ours}\\
Trans-AOPE (\textbf{Ours}) & 71.04 & 67.03 & 68.98 & 60.32 & 44.08 & 50.94 & 72.91 & 66.30 & 69.45 & 73.56 & 64.58 & 68.78 \\
\midrule
\midrule
\textbf{Combined-train setting (Training Sets)}  \\
\midrule
% MiniConGTS (14lap \& 16res) & 74.97 & 71.85 & 73.38 & 73.12 & 63.98 & 68.25 & 94.28 & 91.99 & 93.12 & 77.51 & 77.68 & 77.59 \\
% MiniConGTS (14lap \& 15res) & 74.16 & 70.43 & 72.25 & 72.73 & 67.80 & 70.18 & 68.78 & 71.12 & 69.93 & 73.50 & 76.79 & 75.11 \\
MiniConGTS (14lap \& 14res)
& \cellcolor{lightblue}75.72 & \cellcolor{lightblue}78.20 & \cellcolor{lightblue}76.94
& \cellcolor{lightblue}71.05 & \cellcolor{lightblue}68.64 & \cellcolor{lightblue}69.83
& \cellcolor{lightblue}63.30 & \cellcolor{lightblue}69.90 & \cellcolor{lightblue}66.44
& \cellcolor{lightblue}68.13 & \cellcolor{lightblue}72.54 & \cellcolor{lightblue}70.27 \\

MiniConGTS (14res, 15res, \& 16res)
& \cellcolor{lightyellow}78.68 & \cellcolor{lightyellow}76.78 & \cellcolor{lightyellow}77.72
& \cellcolor{lightyellow}56.86 & \cellcolor{lightyellow}48.31 & \cellcolor{lightyellow}52.23
& \cellcolor{lightyellow}94.54 & \cellcolor{lightyellow}92.48 & \cellcolor{lightyellow}93.50
& \cellcolor{lightyellow}77.65 & \cellcolor{lightyellow}75.22 & \cellcolor{lightyellow}76.42 \\

MiniConGTS (14res, 14lap, 15res, \& 16res)
& \cellcolor{gray}78.22 & \cellcolor{gray}77.11 & \cellcolor{gray}77.66
& \cellcolor{gray}76.32 & \cellcolor{gray}64.19 & \cellcolor{gray}69.74
& \cellcolor{gray}93.35 & \cellcolor{gray}91.99 & \cellcolor{gray}92.67
& \cellcolor{gray}76.27 & \cellcolor{gray}76.79 & \cellcolor{gray}76.53 \\

BARTABSA (14res, 14lap, 15res \& 16res)
& \cellcolor{gray}75.40 & \cellcolor{gray}76.76 & \cellcolor{gray}76.07
& \cellcolor{gray}72.36 & \cellcolor{gray}63.40 & \cellcolor{gray}67.59
& \cellcolor{gray}93.56 & \cellcolor{gray}\textbf{94.43} & \cellcolor{gray}\textbf{93.56}
& \cellcolor{gray}87.06 & \cellcolor{gray}\textbf{87.32} & \cellcolor{gray}87.19 \\

\midrule
\textbf{Ours} \\
Trans-AOPE (14lap \& 14res)
& \cellcolor{lightblue}91.95 & \cellcolor{lightblue}83.24 & \cellcolor{lightblue}87.38
& \cellcolor{lightblue}90.60 & \cellcolor{lightblue}79.88 & \cellcolor{lightblue}84.91
& \cellcolor{lightblue}73.99 & \cellcolor{lightblue}73.19 & \cellcolor{lightblue}73.59
& \cellcolor{lightblue}74.84 & \cellcolor{lightblue}69.94 & \cellcolor{lightblue}72.31 \\

Trans-AOPE (14res, 15res \& 16res)
& \cellcolor{lightyellow}74.34 & \cellcolor{lightyellow}62.29 & \cellcolor{lightyellow}67.78
& \cellcolor{lightyellow}91.03 & \cellcolor{lightyellow}78.11 & \cellcolor{lightyellow}84.07
& \cellcolor{lightyellow}95.66 & \cellcolor{lightyellow}88.04 & \cellcolor{lightyellow}91.69
& \cellcolor{lightyellow}90.63 & \cellcolor{lightyellow}80.65 & \cellcolor{lightyellow}85.35 \\

Trans-AOPE (14res, 14lap, 15res \& 16res)
& \cellcolor{gray}\textbf{92.92} & \cellcolor{gray}\textbf{83.61} & \cellcolor{gray}\textbf{88.02}
& \cellcolor{gray}\textbf{92.20} & \cellcolor{gray}\textbf{80.47} & \cellcolor{gray}\textbf{85.94}
& \cellcolor{gray}\textbf{96.17} & \cellcolor{gray}90.94 & \cellcolor{gray}93.48
& \cellcolor{gray}\textbf{93.75} & \cellcolor{gray}84.82 & \cellcolor{gray}\textbf{89.06} \\
% Trans-AOPE (14res) & 71.04 & 67.03 & 68.98
% & 42.66 & 36.09 & 39.10
% & 71.27 & 71.01 & 71.14
% & 67.34 & 69.35 & 68.33 \\
% Trans-AOPE (14lap) & 66.41 & 47.18 & 55.17
% & 60.32 & 44.08 & 50.94
% & 65.69 & 48.55 & 55.83
% & 64.98 & 49.70 & 56.32 \\
% Trans-AOPE (15res) & 71.85 & 58.11 & 64.25
% & 44.55 & 27.81 & 34.24
% & 72.91 & 66.30 & 69.45
% & 81.96 & 77.08 & 79.45 \\
% Trans-AOPE (16res) & 71.66 & 56.65 & 63.28
% & 41.58 & 24.85 & 31.11
% & 92.13 & 84.78 & 88.30
% & 73.56 & 64.58 & 68.78 \\
% \midrule
% Trans-AOPE (15res \&~16res) & 76.69 & 64.12 & 69.84 & 46.43 & 30.76 & 37.01 & \cellcolor{lightyellow}\textbf{98.81} & \cellcolor{lightyellow}90.58 & \cellcolor{lightyellow}\textbf{94.52} & \cellcolor{lightyellow}\textbf{96.29} & \cellcolor{lightyellow}\textbf{85.12} & \cellcolor{lightyellow}\textbf{90.36} \\

\bottomrule
\end{tabular}
}
\caption{Comparison of different models on multiple datasets for AOPE task. Recall and precision values are omitted where they are not reported. The former best scores are underlined, and current best scores are bold. Highlights are used for analysis. $\diamond$ are retrieved from \citealp{yan-etal-2021-unified}. $\bullet$ is retrieved from \citealp{zhao-etal-2020-spanmlt}, and $\star$ are retrieved from \citealp{sun-etal-2024-minicongts}}
\label{tab:res:aope}
\end{table*}

\section{Results and Analysis}
In this section, we first investigate the effect of adding contrastive loss on performance to determine the optimal training loss configuration (section~\ref{sec:res:contrastive}). Next, we present the results of experiments under different training data configurations, and compare our model with previous methods (section~\ref{sec:main:results}).
We analyze our results in section~\ref{sec:analysis}.
% conducted on various datasets and their combinations, aiming to determine how different training set compositions influence the model’s maximum achievable performance.

\subsection{Effect of Contrastive Loss}\label{sec:res:contrastive}

\begin{figure}[tb!]
\includegraphics[width=0.95\linewidth]{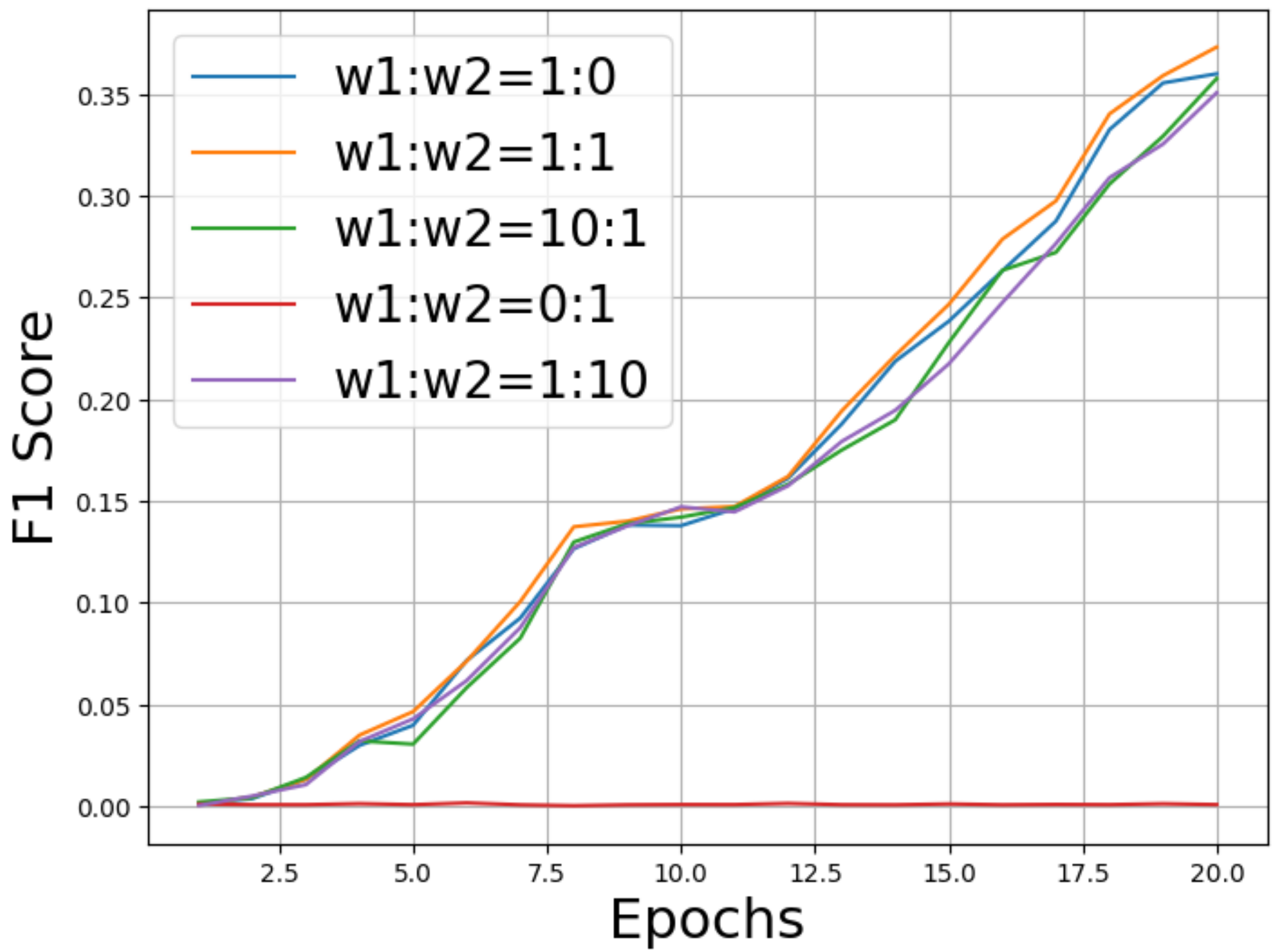}
        \caption{F1 score as a function of training epochs in the combined-train condition for the AOPE task on the 14lap test set, with various loss weight configurations. w1=base loss; w2=contrastive loss. }
        \label{pair}
\end{figure}

To investigate the impact of different weight configurations between the base loss and contrastive loss in Equation~\ref{10}, we evaluated five ratios in \(w_{base}: w_{con}\)—--namely \(1:0\), \(1:1\), \(1:10\), \(0:1\), and \(10:1\)—--with a batch size of 4. The results on 14lap dataset for the AOPE task are shown in Figure~\ref{pair}. Notably, the contrastive-heavy settings where \(w_{base}: w_{con} = 1:1 ~\text{and}~1:10 \) achieved stronger early-stage performance, indicating the initial advantages of adding contrastive optimization.

A similar pattern emerged in the ASTE task, where the base and balanced configurations consistently outperformed the other weight settings, and the balanced option exhibited faster improvements than the base setting. These trends were also observed across additional datasets. Consequently, all subsequent experiments used the balanced configuration \(w_{base} : w_{con} = 1:1\) for optimization.

\subsection{Main Results} \label{sec:main:results}

\paragraph{On AOPE}
First, in the \textbf{in-domain} setting, presented in the upper part of Table~\ref{tab:res:aope}, when models are trained on one dataset only, we observe that our Trans-model (Trans-AOPE) performs worse than previous models.\footnote{Note that the numbers for previous models are retrieved from respective papers that describe the system, and we are assuming they were trained with only in-domain training set, since many of them did not report which training set is used.}

Second, for the \textbf{combined-train} condition, Trans-AOPE has an advantage over MiniConGTS in all combinations of training sets.
What is particularly striking is that, when trained with four training sets combined, Trans-AOPE achieves roughly 20 points or more increase in F1 on all test sets (from 68.98 to 88.02 for 14res, 50.94 to 85.94 for 14lap).  
This is not the case for MiniConGTS, which sees performance drop in two test sets (14res and 14lap), and negligible increase in one (16res), large increase in another (15res: 73.87 to 92.67, but still lower than Trans-AOPE, 93.48). 
This suggests that MiniConGTS is not adaptive and cannot reliably make use of a mixture of training data from two domains and multiple datasets. Similar trends are observed for BARTABSA.

What is more striking is that adding \textit{restaurant} data to the training set (14res and 14lap) improves performance of Trans-AOPE on the \textit{laptop} test set, from 50.94 to 84.91, but not for MiniConGTS, which drops from 73.23 to 69.83. 
Moreover, when trained only on the three restaurant datasets (14res, 15res, and 16res), Trans-AOPE achieves performance comparable to what it attains when the 14lap dataset is included in the training set (84.07 in the yellow row, compared to 84.91 in the blue row). In contrast, MiniConGTS, trained on those same three restaurant datasets, achieves an F1 score of only 52.23. 
% Trans-AOPE, on the other hand, can learn relevant action patterns from the restaurant data and apply them to the 14lap dataset, reaching 84.07 F1—close to the 84.91 or 85.94 F1 obtained when 14lap itself is part of the training set. 
This suggests that Trans-AOPE (unlike previous models) can reliably transfer what it has learned in one dataset to another, even if the data are from a different domain, in the AOPE task.

\begin{table*}[h!]
\centering
\resizebox{\linewidth}{!}{
\begin{tabular}{l|ccc|ccc|ccc|ccc}
\toprule
& \multicolumn{3}{c}{\textbf{14res}} & \multicolumn{3}{c}{\textbf{14lap}} & \multicolumn{3}{c}{\textbf{15res}} & \multicolumn{3}{c}{\textbf{16Res}} \\
\cmidrule(lr){2-4} \cmidrule(lr){5-7} \cmidrule(lr){8-10} \cmidrule(lr){11-13}
\textbf{In-domain setting}& P & R & F1 & P & R & F1 & P & R & F1 & P & R & F1 \\
\midrule
\midrule
Peng-Two-stage \cite{Peng_Xu_Bing_Huang_Lu_Si_2020} $\star$ & 43.24 & 63.66 & 51.46 & 38.87 & 50.38 & 42.87 & 48.07 & 57.51 & 52.32 & 46.96 & 64.24 & 54.21 \\
% GTS \cite{wu2020grid} $\star$ & 67.76 & 67.29 & 67.50 & 57.82 & 51.32 & 54.36 & 62.59 & 57.94 & 60.15 & 66.08 & 66.91 & 67.93 \\
JET-BERT \cite{xu2020position} $\star$ & 70.56 & 55.94 & 62.40 & 55.39 & 43.57 & 51.04 & 64.45 & 51.96 & 57.53 & 70.42 & 58.37 & 63.83 \\
COM-MRC \cite{zhai2022mrc} $\star$ & 75.46 & 68.91 & 72.01 & 58.15 & 60.17 & 61.17 & 68.35 & 61.24 & 64.53 & 71.55 & 71.59 & 71.57 \\
% EMC-GCN \cite{chen2022enhanced} $\star$ & 71.21 & 72.39 & 71.78 & 61.70 & 56.26 & 58.81 & 61.54 & 62.47 & 61.93 & 65.62 & 71.30 & 68.33 \\
% BDTF \cite{chen2022enhanced} $\star$ & 75.53 & 73.24 & 74.35 & 68.94 & 55.97 & 61.74 & 68.76 & 63.71 & 66.12 & 71.44 & 73.13 & 72.27 \\
DGCNAP \cite{li2023dual} $\star$ & 72.90 & 68.69 & 70.72 & 62.02 & 53.79 & 57.57 & 62.23 & 60.21 & 61.19 & 69.75 & 69.44 & 69.58 \\
Triple-MRC \cite{zou2024multi} $\star$ & - & - & 72.45 & - & - & 60.72 & - & - & 62.86 & - & - & 68.65 \\
STAGE-3D \cite{liang2023stage} $\star$         & \underline{78.58} & 69.58          & 73.76          & \underline{71.98} & 53.86          & 61.58          & \underline{73.63} & 57.90          & 64.79          & \underline{76.67} & 70.12          & 73.24          \\
BARTABSA \cite{yan-etal-2021-unified} $\diamond$ & 65.52 & 64.99 & 65.25 & 61.41 & 56.19 & 58.69 & 59.14 & 59.38 & 59.26 & 66.60 & 68.68 & 67.62 \\
MiniConGTS \cite{sun-etal-2024-minicongts} $\star$ & 76.10          & \underline{75.08} & \underline{75.59} & 66.82          & \underline{60.68} & \underline{63.61} & 66.50          & \underline{63.86} & \underline{65.15} & 75.52          & \underline{74.14} & \underline{74.83} \\
\midrule
Trans-ASTE (\textbf{Ours}) & 65.46 & 59.38 & 62.27 & 53.44 & 41.42 & 46.67 & 66.28 & 61.96 & 64.04 & 64.82 & 59.23 & 61.90 \\
\midrule
\midrule
\textbf{Combined-train setting (Training Sets)} \\
\midrule
% MiniConGTS (14lap \& 16res) & 70.97 & 68.02 & 69.46 & 66.10 & 57.84 & 61.69 & 92.29 & 90.05 & 91.15 & 73.05 & 73.21 & 73.13 \\

% MiniConGTS (14lap \& 15res) & 70.93 & 67.36 & 69.10 & 65.68 & 61.23 & 63.38 & 61.27 & 63.35 & 62.29 & 69.66 & 72.77 & 71.18 \\

% Trans-ASTE (14res) & 65.46 & 59.38 & 62.27
% & 40.60 & 31.95 & 35.76
% & 68.28 & 66.30 & 67.28
% & 63.50 & 63.69 & 63.60 \\
% Trans-ASTE (14lap) & 57.47 & 46.27 & 51.26
% & 53.44 & 41.42 & 46.67
% & 62.99 & 51.81 & 56.86
% & 55.05 & 47.02 & 50.72 \\
% Trans-ASTE (15res) & 65.45 & 55.56 & 60.10
% & 36.65 & 23.96 & 28.98
% & 66.28 & 61.96 & 64.04
% & 75.94 & 72.32 & 74.09 \\
% Trans-ASTE (16res) & 62.90 & 53.73 & 57.96
% & 33.47 & 23.37 & 27.53
% & 83.77 & 80.43 & 82.07
% & 64.82 & 59.23 & 61.90 \\

MiniConGTS (14lap \& 14res)
& \cellcolor{lightblue}72.11 & \cellcolor{lightblue}74.48 & \cellcolor{lightblue}73.28
& \cellcolor{lightblue}62.50 & \cellcolor{lightblue}60.38 & \cellcolor{lightblue}61.42
& \cellcolor{lightblue}56.48 & \cellcolor{lightblue}62.38 & \cellcolor{lightblue}59.28
& \cellcolor{lightblue}64.57 & \cellcolor{lightblue}68.75 & \cellcolor{lightblue}66.59 \\

MiniConGTS (14res, 15res, \& 16res)
& \cellcolor{lightyellow}74.64 & \cellcolor{lightyellow}72.84 & \cellcolor{lightyellow}73.73
& \cellcolor{lightyellow}52.12 & \cellcolor{lightyellow}44.28 & \cellcolor{lightyellow}47.88
& \cellcolor{lightyellow}92.31 & \cellcolor{lightyellow}90.29 & \cellcolor{lightyellow}91.29
& \cellcolor{lightyellow}74.42 & \cellcolor{lightyellow}72.10 & \cellcolor{lightyellow}73.24 \\

MiniConGTS (14res, 14lap, 15res \& 16res)
& \cellcolor{gray}73.89 & \cellcolor{gray}72.84 & \cellcolor{gray}73.36
& \cellcolor{gray}68.26 & \cellcolor{gray}57.42 & \cellcolor{gray}62.37
& \cellcolor{gray}90.64 & \cellcolor{gray}89.32 & \cellcolor{gray}89.98
& \cellcolor{gray}72.73 & \cellcolor{gray}73.21 & \cellcolor{gray}72.97 \\

BARTABSA (14res, 14lap, 15res \& 16res)
& \cellcolor{gray}71.05 & \cellcolor{gray}72.33 & \cellcolor{gray}71.68
& \cellcolor{gray}63.66 & \cellcolor{gray}56.01 & \cellcolor{gray}59.59
& \cellcolor{gray}91.30 & \cellcolor{gray}\textbf{92.99} & \cellcolor{gray}92.13
& \cellcolor{gray}83.82 & \cellcolor{gray}\textbf{84.07} & \cellcolor{gray}83.95 \\

\midrule
\textbf{Ours} \\

Trans-ASTE (14lap \&14res)
& \cellcolor{lightblue}88.05 & \cellcolor{lightblue}80.51 & \cellcolor{lightblue}84.11
& \cellcolor{lightblue}\textbf{88.11} & \cellcolor{lightblue}74.56 & \cellcolor{lightblue}80.77
& \cellcolor{lightblue}69.55 & \cellcolor{lightblue}67.03 & \cellcolor{lightblue}68.27
& \cellcolor{lightblue}64.53 & \cellcolor{lightblue}62.80 & \cellcolor{lightblue}63.65 \\

Trans-ASTE (14res, 15res \& 16res)
& \cellcolor{lightyellow}73.53 & \cellcolor{lightyellow}61.74 & \cellcolor{lightyellow}67.12
& \cellcolor{lightyellow}42.25 & \cellcolor{lightyellow}32.25 & \cellcolor{lightyellow}36.58
& \cellcolor{lightyellow}89.89 & \cellcolor{lightyellow}86.96 & \cellcolor{lightyellow}88.40
& \cellcolor{lightyellow}81.19 & \cellcolor{lightyellow}77.08 & \cellcolor{lightyellow}79.08 \\

Trans-ASTE (14res, 14lap, 15res \& 16res)
& \cellcolor{gray}\textbf{89.56} & \cellcolor{gray}\textbf{81.24} & \cellcolor{gray}\textbf{85.20}
& \cellcolor{gray}86.87 & \cellcolor{gray}\textbf{76.33} & \cellcolor{gray}\textbf{81.26}
& \cellcolor{gray}\textbf{94.34} & \cellcolor{gray}90.58 & \cellcolor{gray}\textbf{92.42}
& \cellcolor{gray}\textbf{90.03} & \cellcolor{gray}83.33 & \cellcolor{gray}\textbf{86.55} \\
% Trans-ASTE (15res \&16res)  & 70.05 & 55.37 & 61.85 & 41.58 & 24.85 & 31.11 & \cellcolor{lightyellow}\textbf{98.40} & \cellcolor{lightyellow}89.49 & \cellcolor{lightyellow}\textbf{93.73} & \cellcolor{lightyellow}\textbf{93.60} & \cellcolor{lightyellow}82.74 & \cellcolor{lightyellow}87.83 \\

\bottomrule
\end{tabular}
}
\caption{Comparison of different models on multiple datasets for ASTE task. Recall and precision values are omitted where they are not reported. Highlights are used for analysis. The former best scores are underlined, and current best scores are bold. $\star$ are retrieved from \citealp{sun-etal-2024-minicongts}, and  $\diamond$ is retrieved from \citealp{yan-etal-2021-unified}.}
\label{tab:res:aste}
\end{table*}

\paragraph{On ASTE} 
From Table~\ref{tab:res:aste} we observe similar patterns as AOPE, where our Trans-model demonstrates worse performance in the \textbf{in-domain} setting but superior performance under the \textbf{combined-train} setting. 
In the \textbf{in-domain} setting, Trans-ASTE is on par with previous models in 15res, but behind MiniConGTS on the other three test sets. 
However, under the \textbf{combined-train} condition, we again observe that our Trans-model benefits from the mixture of training data.
For instance, while testing on 14res, the performance gains a 20 points boost when adding the training data of 14lap to 14res (62.27 to 85.20). On the contrary, for MiniConGTS, we see a drop in performance from 75.59 to 73.28. For BARTABSA, there is a moderate improvement (65.25 to 71.68). 
When four training sets are combined, Trans-ASTE shows superior performance in all test sets except 15res, usually by a large margin (gray rows in Table~\ref{tab:res:aste}). 
This is especially true for 14lap, in which Trans-ASTE is the only model benefiting tremendously from adding restaurant data (performance increases from 46.67 to 81.26). 

The only pattern different from Trans-AOPE is that when training on three restaurant datasets and testing on 14lap (yellow row), Trans-ASTE has very low performance of 36.58 F1, much lower than the previous 84.07 F1 in Trans-AOPE. 
% Note that this scenario essentially involves cross-domain transfer with sentiment tagging. 
We believe that this is due to the inability of the sentiment tagging system to transfer from one domain to another, rather than the issue of the aspect–opinion pair extraction of our transition-based pipeline. 
We thus find that Trans-model demonstrates stronger transfer performance in the AOPE task compared to ASTE, possibly because AOPE only requires predicting actions to extract aspect–opinion pairs. In contrast, ASTE requires both identifying these pairs and assigning a sentiment tag to form triplets, and sentiment tags tend to be more domain-specific.

% \textit{BARTABSA} and \textit{MiniConGTS} were trained on OmniTrain (4-in-1) datasets, demonstrating substantial gains on the 15res and 16res datasets, with APGs of up to 26.98\% and 9.12\%, respectively. While \textit{BARTABSA} showed some improvement on the 14res and 14lap datasets, \textit{MiniConGTS} suffered a performance drop on these benchmarks. Notably, 14res, 14lap, and 16res remain challenging for the ASTE task. When trained on separate datasets, \textit{Trans-ASTE} displayed a similar pattern to that observed in the AOPE task: substantial improvement on 15res but lower performance on 14res, 14lap 16res. To address this, we combined two datasets for training, which yielded APGs of 10.75\%, 18.4\%, 1.6\%, and 4.94\% on 14res, 14lap, 15res, and 16res, respectively, compared to a baseline trained on the fused dataset. Finally, our model in the combined-train setting achieved slightly lower performance comparing to models trained on 2-in-1 datasets highlighted in blue and yellow for 14res\&14lap and 15res\&16res.

\subsection{When and Why is Trans-model Better?} \label{sec:analysis}

\textbf{When: Trans-model is better with combined training sets and in multi-domain generalization.}
From the results above, it seems clear that the Trans-model in the two tasks are better when trained on multiple datasets combined, rather than one dataset alone. 
Our results also suggest that the added training data do not have to be in the same domain: when trained on 14lap \& 14res and tested on either 14lap or 14res, the F1 score is at least about 10 percentage points better than trained on only one of the two datasets.
It shows that our Trans-model is capable of learning from multiple domains, and seems to be able to transfer its knowledge from the laptop domain to the restaurant domain and vice versa, unlike MiniConGTS, which seems to be more sensitive to the domain of the data. 
However, further research is need to better understand the discrepancy between Trans-AOPE and Trans-ASTE models in cross-domain transfer, and improve the cross-domain transfer ability of the sentiment tagger. 

% \orange{TODO: Secondly, replacing individual training with a combined training approach does not necessarily guarantee improved performance for every model. For instance, when \textit{MiniConGTS} is trained on the OmniTrain dataset, it experiences slight performance drops on 14res, 14lap, and 16res. Likewise, when \textit{Trans-AOPE} and \textit{Trans-ASTE} are trained on the OmniTrain dataset, both show clear declines compared to their 2-in-1 counterparts across all four datasets. A plausible explanation is overfitting to restaurant-related features from the relatively simpler 15res dataset, which may overshadow patterns in other datasets. In general, merging multiple datasets for “all-in-one” training does not always yield gains for state-of-the-art models. Nevertheless, our model, when trained on fused datasets, consistently achieves improvements across multiple benchmarks, with particularly notable increases on 14res, 14lap, and 16res. These findings highlight the need for careful consideration of dataset combinations. In-domain setting may not always be the best strategy for every model.
% }

\textbf{Why: Trans-model can learn more actions in data from diverse domains.}
We believe that the proposed trans-models excel when trained on multiple datasets and show considerable generalization ability for two main reasons. First, by predicting actions instead of tokens, it avoids token-level biases and prevents overfitting on token-level patterns specific to restaurant datasets. Additionally, trans-models perform pair extraction after the aspect–opinion relationship is established, which enables the model to capture contextual relationships more effectively. These design decisions work together to significantly enhance the overall effectiveness and robustness of the trans-models.

% \orange{TODO:
% These results yield several noteworthy findings. First, 15res and 16res appear to be relatively simpler datasets, while 14res and 14lap pose greater challenges for both baseline models, even when the OmniTrain dataset is used. One possible explanation is that 14res contains a larger volume of data, and 14lap comprises laptop reviews rather than restaurant reviews, making it more difficult for all models to capture underlying patterns. In particular, for the 14lap dataset, aspect entities differ considerably from those in the restaurant datasets, causing the baseline model to exhibit a slight performance decline when using the fused dataset. In practice, having mixed, unbalanced, and unsorted data is more common. Consequently, an optimal model should account for the complexity of real-world language features and demonstrate robustness against these disadvantages. We show that the proposed transitional model is a better option, as it has the capacity to learn fundamental language patterns regardless of the aspect categories.
% }

\section{Conclusion and Future Work}
In this paper, we present an efficient transition pipeline for the extraction of aspects-opinion pairs with linear time complexity \(O(n)\), enhanced by a contrastive-based optimization method. This approach obviates the need to directly identify and extract individual tokens, thereby mitigating token-level bias. It can be trained on a combination of diverse datasets that offers the most comprehensive coverage of the actions needed, resulting in significant performance improvements across various datasets. Specifically, training our model on a well-covered fused dataset enables it to learn robust action patterns, leading to superior performance on all datasets. Our model surpasses retrained baseline models on the same fused dataset, establishing new state-of-the-art results for both AOPE and ASTE tasks.

As transition-based methods have remained relatively less explored in sentiment-related tasks, we believe our work shows a promising direction to employ such methods in aspect-based sentiment analysis. 
Future work can further examine the potential of transition-based models in other sentiment analysis tasks, as well as the generalization ability of these models in situations of multi-domain data. 
It is also important to better understand the cross-domain and multi-domain generalization ability of transition models, by experimentation on more domains, since only two domains are involved in this work.  

\section*{Limitation}
Although the Trans-model demonstrates robust generalization capability, its reliance on larger datasets to effectively learn action patterns remains a notable limitation of the transition-based pipeline. This issue is evident in our results: while it is not necessary to use the \textit{same} dataset for both training and testing, the model performs better when trained on blended datasets rather than on a single, limited one. 
Consequently, if the training data lack sufficient action patterns, the model’s ability to handle nuanced or previously unseen contexts can be significantly compromised. 
These findings underscore the importance of training on a combined or broader dataset to enhance the model’s overall effectiveness.

\bibliography{acl_latex}

\end{document}

%% file: new-intro.tex
\section{Introduction}
% Sentiment analysis, the computational study of opinions, emotions, and attitudes in text, has emerged as a vital research area with applications spanning from business to social sciences \cite{Pang2002, Liu2012, Cambria2016, wang2025dlf}. 
Aspect-Based Sentiment Analysis (ABSA) is a fine-grained sentiment analysis task that identifies specific aspects in text and analyzes the sentiments linked to them \cite{Hu2004,Liu2012,wang2025dlf}. 
As shown in Figure~\ref{d}, ABSA involves subtasks such as Aspect Extraction (AE) and Opinion Extraction (OE)---identifying mentioned aspects and their related opinions, or the combination---Aspect-Opinion Pair Extraction.
Once the aspect and opinion have been extracted, a sentiment is usually computed, and this more complicated task is often referred to as Aspect-Sentiment Triplet Extraction (ASTE). 

\input{first-fig}

For instance, given the sentence: ``\textit{Gourmet food is delicious. Good service, but not so welcoming}'', AE identifies \textbf{\textcolor{blueish}{gourmet food}} and \textbf{\textcolor{blueish}{service}} as aspects, while OE extracts \textbf{\textcolor{orange}{delicious}}, \textbf{\textcolor{orange}{good}}, and \textbf{\textcolor{orange}{not so welcoming}} as opinions. 
These outputs are then combined to form aspect-opinion pairs, with a separate sentiment tagging system assigning polarities to create triplets~\cite{jiang2023semanticallyenhanceddualencoder, 10.1007/978-3-031-25198-6_8, chakraborty2024aspectopiniontermextraction}.

% It involves subtasks such as: \textbf{\textit{Aspect Extraction (AE)}}: Identifying mentioned aspects. \textbf{\textit{Opinion Extraction (OE)}}: Extracting opinion terms related to aspects. \textbf{\textit{Aspect-Opinion Pair Extraction (AOPE)}}: Extracting aspect-opinion pairs. \textbf{\textit{Aspect-Sentiment Triplet Extraction (ASTE)}}: Identifying aspect, opinion, and sentiment polarity triplets.

% Classic ABSA approaches often use a pipeline architecture, performing the subtasks separately. 
% For instance, given the sentence: ``\textit{Gourmet food is delicious. Good service, but not so welcoming}'', AE identifies \textbf{\textcolor{blueish}{gourmet food}} and \textbf{\textcolor{blueish}{service}} as aspects, while OE extracts \textbf{\textcolor{orange}{delicious}}, \textbf{\textcolor{orange}{good}}, and \textbf{\textcolor{orange}{not so welcoming}} as opinions. 
% These outputs are then combined to form aspect-opinion pairs, with a separate sentiment tagging system assigning polarities to create triplets. 
% Past studies have generally achieved satisfactory performance 
% \fly{No need to write this score} 
% on aspect and opinion extraction \cite{jiang2023semanticallyenhanceddualencoder, 10.1007/978-3-031-25198-6_8, chakraborty2024aspectopiniontermextraction}. 

% Adding the according sentiment tags with pair, 
ASTE is the most integrated task for aspect-based sentiment analysis, for which diverse models leveraging various methodologies have been developed, including pipeline-based approach~\citep{Peng_Xu_Bing_Huang_Lu_Si_2020}, sequence-to-sequence  method~\citep{yan-etal-2021-unified}, sequence-tagging method~\citep{wu2020grid,xu2020position},  to name just a few. 
Despite these efforts and growing interests, the accuracy of recent models remains suboptimal, with the best systems scoring 60\% or 70\%~\cite{sun-etal-2024-minicongts}. 
There are two key challenges that hinder performance:
(1) \textbf{Disconnected Aspect-Opinion Extraction}: Opinions are often extracted independently from their corresponding aspects~\cite{liang2023stage, sun-etal-2024-minicongts}. While positional relationships can be added as an auxiliary factor to assist pair extraction \cite{liu2022mrce, 10.1007/978-3-031-25198-6_8}, this approach loses critical contextual information by treating aspects and opinions as separate entities. This limits the effectiveness of many token-based extraction methods. 
(2) \textbf{High time complexity with longer sequences}: Methods using 2D matrix tagging \cite{liang2023stage, sun-etal-2024-minicongts} to capture relationships between tokens face significant increases in time complexity as the length of the token sequence increases. 
This computational burden restricts their scalability, especially for longer texts in practical applications.

To address these two challenges, we present the first transition-based AOPE system named \textbf{Trans-AOPE} that (1) extracts the Aspect and the Opinion at the same time, and (2) has a time complexity of $O(n)$. 
% Building on the proposed transition-based framework,
We also introduce a contrastive-augmented optimization method to enhance model efficiency. 
We conduct experiments on 4 commonly used ABAS datasets, and compare our system with previous models.
Our results show that \textbf{Trans-model} achieves state-of-the-art performance on all datasets we tested.
We conduct comprehensive ablation studies to evaluate the contribution of optimization components and perform extensive training on various datasets to identify precisely where our model and baselines derive their learning. Our contributions are:

\begin{itemize}[topsep=0pt, itemsep=0pt, parsep=0pt]
\item We propose the first transition-based model that extracts aspect-opinion \textit{pairs} based on relational aspects, rather than using relational factors as supplementary references or confirmation, with linear time complexity.
% transition pipelines \textbf{Trans-AOPE} and \textbf{Trans-ASTE} with linear time complexity, the first to extract aspect-opinion pairs based on relational aspects rather than using relational factors as supplementary references or confirmation.
\item We experiment with a contrastive-augmented optimization method and find that balanced weighting yields faster, more stable improvements, emerging as the optimal training configuration.
% Building on the proposed transitional framework, we introduce a contrastive-augmented optimization method to improve model efficiency. Our experiments explore how contrastive learning impacts convergence and performance of trans-models.
\item We explore various training strategies and show that our proposed method achieves optimal performance on four datasets when trained on combined training sets, with better cross-dataset generalization.
% Lastly, we also explored various training strategies. The proposed models, Trans-AOPE and Trans-ASTE, achieve optimal performance on four datasets in just one training and demonstrate better cross-dataset generalization.
\end{itemize}

%% file: first-fig.tex
\begin{figure}[t]
\begin{center}
\includegraphics[width=\linewidth]{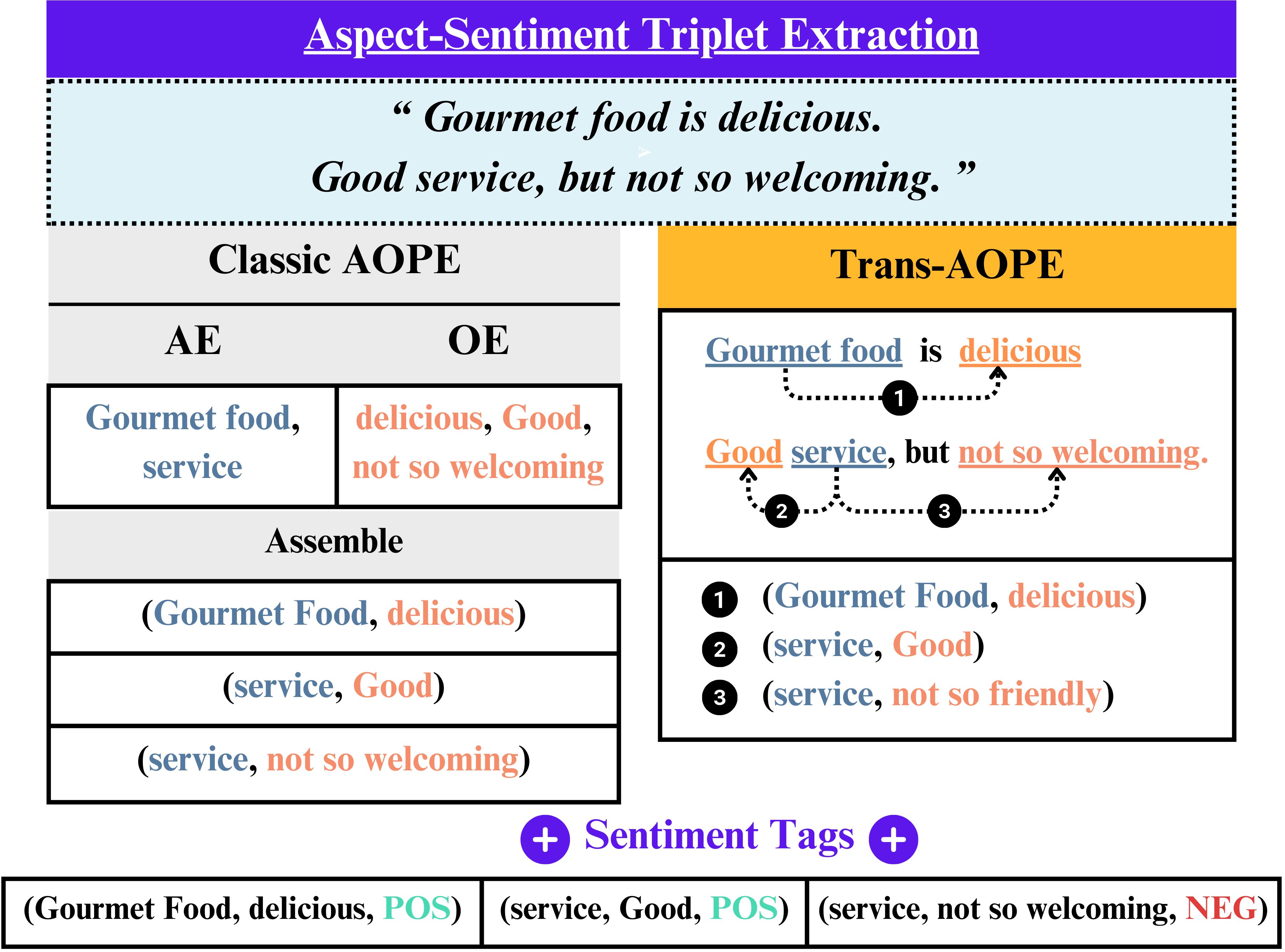}
\caption{
% For the sentence ``Gourmet food is delicious. Good service, but not so welcoming'', this example demonstrates 
Demonstration of the processing steps in both classic and transitional methods for extracting aspect-opinion pairs and tagging sentiment polarity. 
Importantly, our proposed transitional method predicts transition actions, and performs pair extraction after the aspect–opinion relationship has been established, allowing the model to capture contextual relationships more effectively.
AE stands for Aspect Extraction, OE for Opinion Extraction. \textcolor{customgreen}{POS} and \textcolor{rr}{NEG} represent positive and negative sentiments respectively.
}
\label{d}
\end{center}
\end{figure}